\documentclass{article}

\PassOptionsToPackage{numbers, compress}{natbib}

\usepackage[final]{neurips_2023}

\usepackage[utf8]{inputenc} 
\usepackage[T1]{fontenc}    
\usepackage{hyperref}       
\usepackage{url}            
\usepackage{booktabs}       
\usepackage{amsfonts}       
\usepackage{nicefrac}       
\usepackage{microtype}      
\usepackage{xcolor}         
\usepackage{comment}
\usepackage{graphicx}
\usepackage{array}

\newcommand{\customwidth}{\textwidth}

\title{LDM3D-VR: Latent Diffusion Model for 3D VR}

\newcommand{\tab}{\hspace*{2em}}
\author{%
  Gabriela Ben Melech Stan\tab 
  Diana Wofk\tab 
  Estelle Aflalo\tab \\
  \textbf{Shao-Yen Tseng\tab 
  Zhipeng Cai\tab 
  Michael Paulitsch\tab 
  Vasudev Lal}\\
  Intel Labs\\
  \tt\small{\{gabriela.ben.melech.stan, diana.wofk, estelle.aflalo,} \\ \tt\small{shao-yen.tseng, zhipeng.cai, michael.paulitsch, vasudev.lal\}@intel.com} \\
}

\begin{document}

\maketitle

\begin{abstract}

Latent diffusion models have proven to be state-of-the-art in the creation and manipulation of visual outputs. However, as far as we know, the generation of depth maps jointly with RGB is still limited. We introduce LDM3D-VR, a suite of diffusion models targeting virtual reality development that includes LDM3D-pano and LDM3D-SR. These models enable the generation of panoramic RGBD based on textual prompts and the upscaling of low-resolution inputs to high-resolution RGBD, respectively. Our models are fine-tuned from existing pretrained models on datasets containing panoramic/high-resolution RGB images, depth maps and captions. Both models are evaluated in comparison to existing related methods. 
\end{abstract}

\section{Introduction}

Diffusion models have brought about a shift in content creation by offering accessible models that generate RGB images from text prompts or produce high-resolution RGB images from low-resolution inputs. However, the generation of depth maps jointly with RGB images, often required for virtual reality (VR) content development, still presents a challenge. Panoramas created using conventional image stitching algorithms exhibit certain drawbacks, such as artifacts and irregular shapes.

LDM3D-VR builds upon the Latent Diffusion Model for 3D (LDM3D) \cite{ldm3d} and explores RGBD generation for panoramic views. We also create a super-resolution model based on LDM3D that jointly upscales an image alongside its corresponding depth map.
To summarize, our contributions are the following: (i) we introduce LDM3D-pano, which addresses the challenge of generating panorama views jointly with their corresponding depth maps based on a text prompt (ii) we introduce LDM3D-SR that performs x4 upscaling and recovers high-resolution RGB and depth maps from low-resolution inputs (iii) we showcase this work through a demo accessible at ~\url{https://huggingface.co/spaces/Intel/ldm3d}. LDM3D-pano and LDM3d-SR are available at~\url{https://huggingface.co/Intel}.

\section{Related work} 

\textbf{Text-to-panorama.} Text-to-panorama is an important task for creating VR environments. 
Early approaches achieve this goal through Generative Adversarial Networks (GANs)~\cite{lin2021infinitygan, chen2022text2light}. 
Recent advances in diffusion models have improved training stability and model generalization capacity compared to GAN, enabling text-to-panorama works ~\cite{bar2023multidiffusion, tang2023mvdiffusion, blockade_lab}. Some of these methods~\cite{tang2023mvdiffusion, bar2023multidiffusion} only cover the left right rotations of panorama, i.e., without top down views. Others~\cite{blockade_lab} cannot generate realistic panoramas due to the lack of training data. We propose LDM3D-pano, a novel diffusion-based approach capable of producing, from an input text prompt, a realistic RGB panorama and its corresponding panorama depth map. 

\textbf{Super-resolution for images.} Learning-based super-resolution (SR) has been extensively studied in the past decade.
Following \cite{dong2015image}, initial methods adopted convolution neural networks (CNN) and proposed various methods to improve reconstruction quality \cite{kim2016deeply,kim2016accurate,zhang2018residual,lim2017enhanced, zhang2018image}.
GANs were later introduced, which led to higher fidelity SR images \cite{ledig2017photo,wang2018esrgan,wang2021real}.
Subsequent approaches then improved SR performance through the use of attention \cite{zhang2018image,dai2019second,niu2020single} and transformer-based architectures \cite{liang2021swinir,mei2021image,wang2022uformer,Lu_2022_CVPR,Chen_2023_CVPR}.
Most recently, denoising diffusion probabilistic models \cite{ho2020denoising} have demonstrated proficiency in image generation \cite{rombach2022high,saharia2022photorealistic} as well as image upscaling~\cite{saharia2022image,shang2023resdiff,Gao_2023_CVPR}.

\textbf{Super-resolution for depth.} Enhancing the resolution of a depth map is also a widely-studied problem. Naive pixel-level interpolations often yield noisy floating points at object boundaries. Learning-based approaches~\cite{xie2015edge, li2020depth, guo2022depth, yang2022codon, zhao2022discrete, ariav2023fully} have emerged as promising alternatives. \\
In this work, we propose LDM3D-SR, a latent diffusion-based super-resolution model that can enhance the resolution of RGB and depth maps within the same architecture.

\section{Methodology}

\subsection{LDM3D-pano}

LDM3D-pano extends LDM3D~\cite{ldm3d} to panoramic image generation. Key changes to the architecture include adjustments to the first and last Conv2d layers of the KL-autoencoder \cite{esser2021taming}, enabling it to process a 4-channel input consisting of RGB concatenated with a single-channel depth map; we denote this model as LDM3D-4c. The employed diffusion model is based on U-Net \cite{ronneberger2015unet} operating in a 64x128x4 latent space, following \cite{rombach2022highresolution}, with the incorporation of a CLIP text encoder \cite{radford2021learning} for text conditioning through cross-attention mapping on the U-Net layers. 

\label{ldm3d-pano-finetune}
We adopt a two-stage fine-tuning procedure, following~\cite{rombach2022high,ldm3d}. We first fine-tune the refined version of the KL-autoencoder in LDM3D-4c, using roughly 10k samples of size 256x256 sourced from LAION-400M~\cite{laion400M}, with depth map labels produced using the DPT-BEiT-L-512~\cite{birkl2023midas}.

Subsequently, the U-Net backbone is fine-tuned based on Stable Diffusion (SD) v1.5 \cite{rombach2022high}, employing a subset of LAION Aesthetics 6+ \cite{schuhmann2022laion5b} consisting of nearly 20k tuples (captions, 512x512-sized images and depth maps produced using DPT-BEiT-L-512~\cite{birkl2023midas}).
We further fine-tune the U-Net on our panoramic dataset, comprised of High Dynamic Range (HDR) images--originally in 4k resolution--sourced from \cite{polyhaven}, \cite{ihdri}. These HDRIs are augmented into 512x1024 panoramic images utilizing the methodology from \cite{chen2023text2light}, producing 7828 training images and 322 validation images. Panoramic depth maps labels at 512x1024 resolution are produced using DPT-BEiT-L-512~\cite{birkl2023midas})
Captions are generated using BLIP-2 \cite{li2023blip2}. Of the resulting captions, \(\sim70\% \) start with "360 view of" and \(\sim4\% \) with "panoramic view of," while the remaining captions do not feature a panorama-related mention.

\subsection{LDM3D-SR}

LDM3D-SR specializes in super-resolution, utilizing the KL-AE previously developed for LDM3D-4c~\ref{ldm3d-pano-finetune} to now encode low-resolution (LR) images into a 64x64x4 dimensional latent space. The diffusion model used here is an adapted version of the U-Net referenced in \ref{ldm3d-pano-finetune}, now modified to have an 8-channel input. This change enables conditioning on LR latent via concatenation to the high-resolution (HR) latent during training, and to noise during inference. Text conditioning is also facilitated using cross attention with a CLIP text encoder.

\label{ldm3d-hr-finetuned}
We finetune the U-Net in LDM3D-SR from SD-superres~\cite{superres}. Training data consists of HR and LR sets with 261,045 samples each. For HR samples, we use a subset of LAION Aesthetics 6+ with tuples (captions, 512x512-sized images, and depth maps from DPT-BEiT-L-512~\cite{birkl2023midas}). LR images are generated using a lightweight BSR-image-degradation method, introduced in \cite{rombach2022high} applied to the HR image. We explored three methods for generating LR depth maps: performing depth estimation on the LR depth maps (LDM3D-SR-D), utilizing the original HR depth map for LR conditioning (LDM3D-SR-O), and applying bicubic degradation to the depth map (LDM3D-SR-B). 

\section{Results}

\subsection{Panoramic RGBD generation}

We evaluate text-to-pano RGBD generation using the validation set of our dataset (see ~\ref{ldm3d-pano-finetune}).

\textbf{Image evaluation.} We compare LDM3D-pano to Text2light LDR~\cite{chen2023text2light}, a model that creates a text-driven Low Dynamic Range panorama using a hierarchical approach for detail rendering, where global text-scene alignment is followed by a local sampler to facilitate patch-based panorama synthesis.

\begin{figure}[!htb]
\begin{minipage}{.7\linewidth}
      \renewcommand{\customwidth}{0.25\linewidth}
      \setlength{\tabcolsep}{1pt}
      \centering
      \begin{tabular}{c c c c}
        \scriptsize{Text2light~\cite{chen2023text2light} Image} & 
        \scriptsize{\textbf{LDM3D-pano Image}} & 
        \scriptsize{\textbf{LDM3D-pano Depth}}  & 
        \scriptsize{Joint\_3D60~\cite{yun2022improving} Baseline} 
        \\
        \vspace{-1pt}
        \includegraphics[width=\customwidth, keepaspectratio]{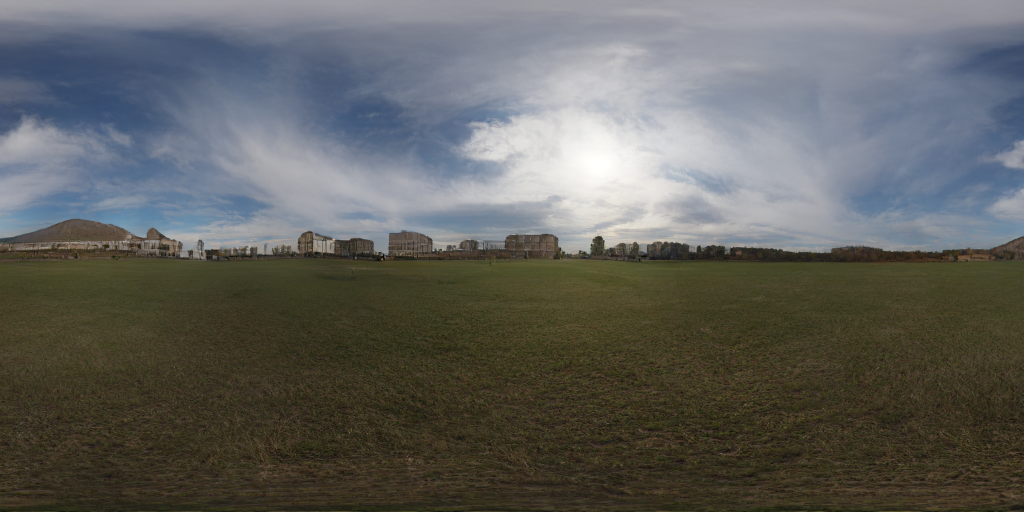} &	    
        \includegraphics[width=\customwidth, keepaspectratio]{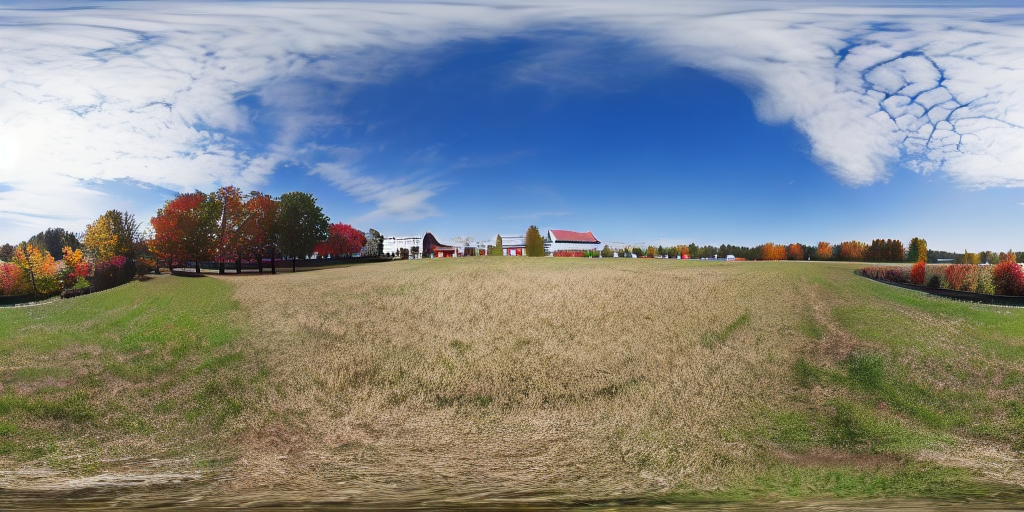} &
        \includegraphics[width=\customwidth, keepaspectratio]{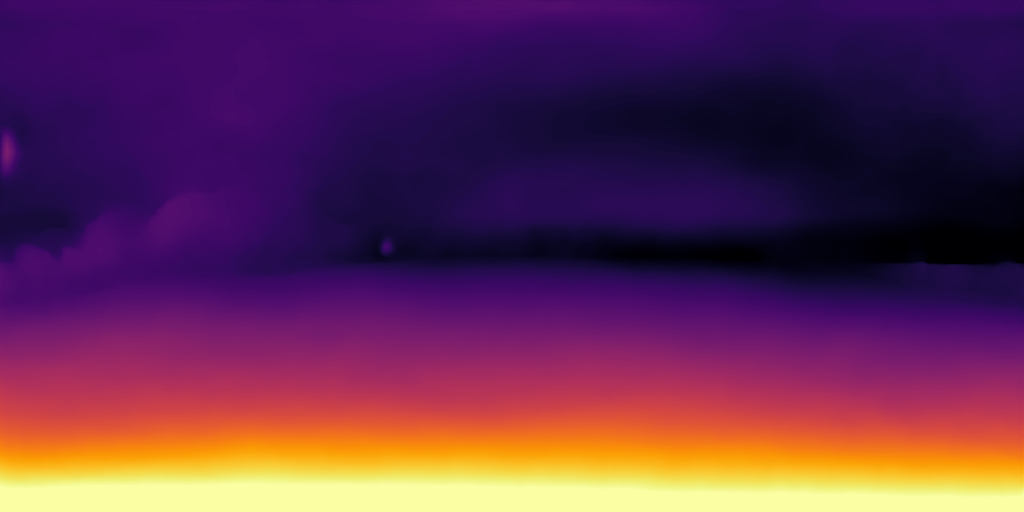} &
    	\includegraphics[width=\customwidth, keepaspectratio]{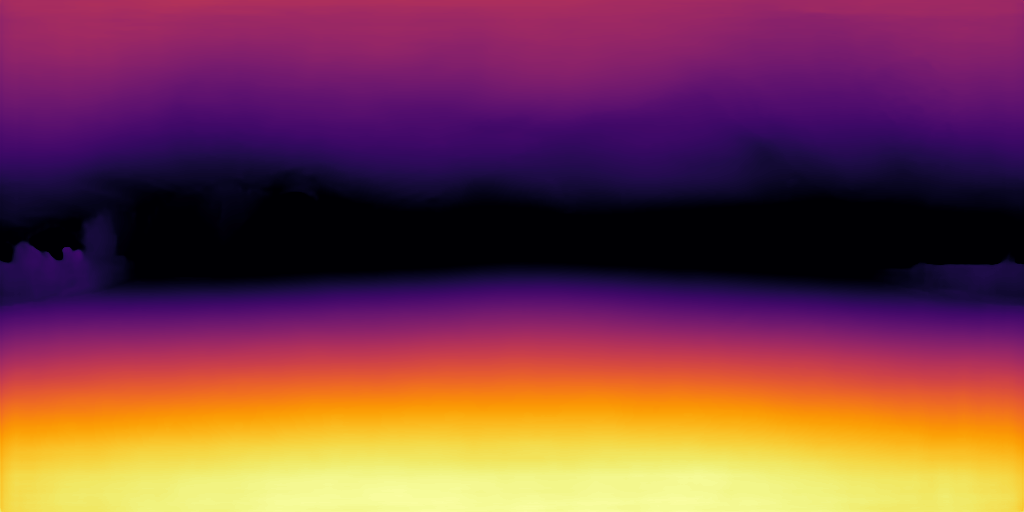}
        \\
        \includegraphics[width=\customwidth, keepaspectratio]{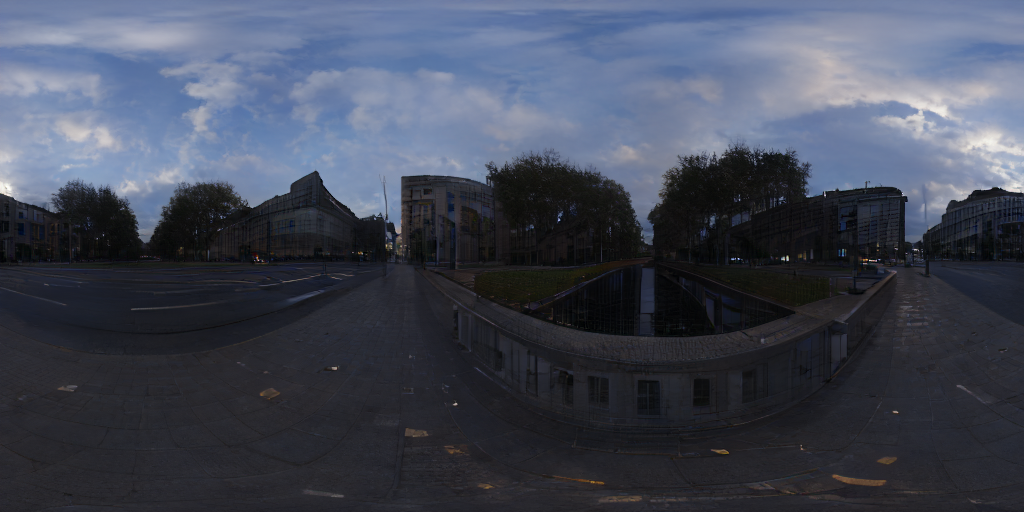} &	     
    	\includegraphics[width=\customwidth, keepaspectratio]{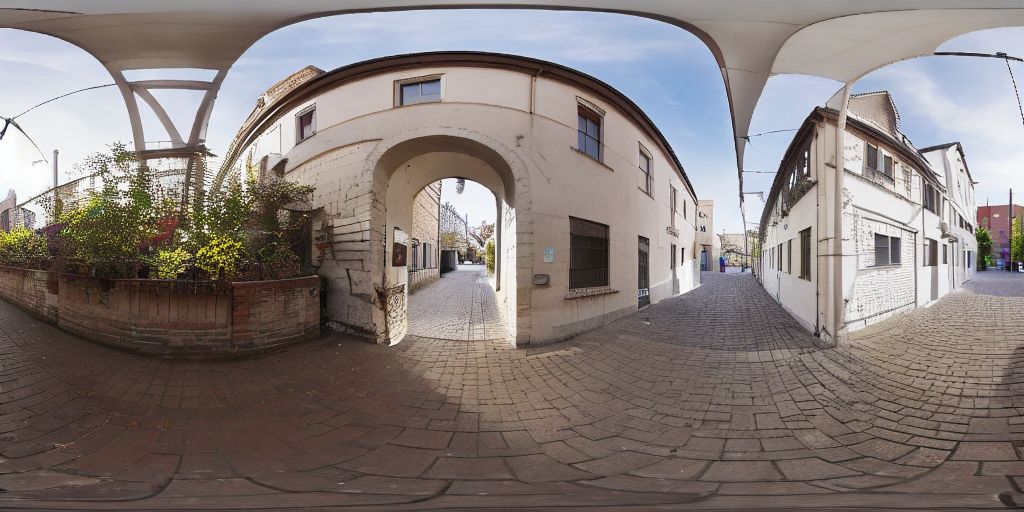} &
    	\includegraphics[width=\customwidth, keepaspectratio]{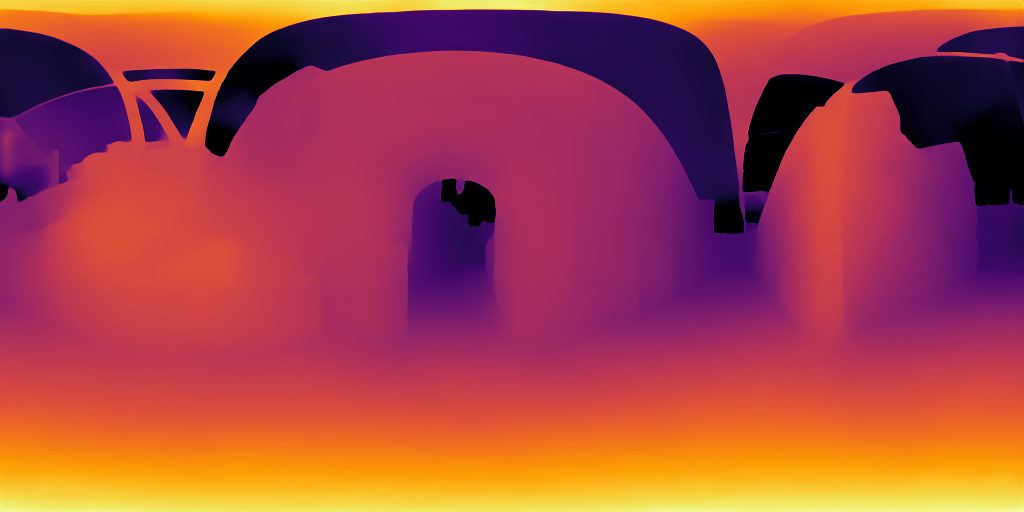} &
    	\includegraphics[width=\customwidth, keepaspectratio]{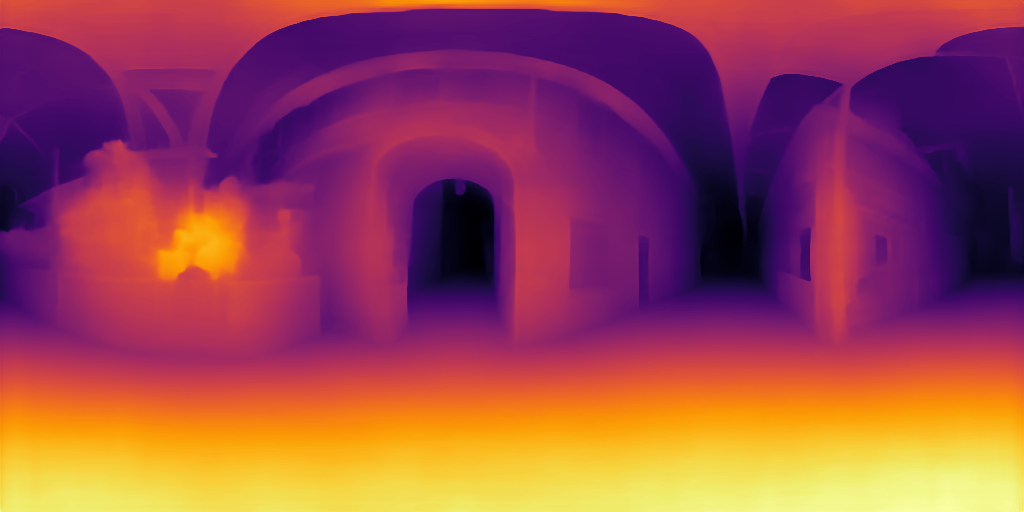}
        \\
        \includegraphics[width=\customwidth, keepaspectratio]{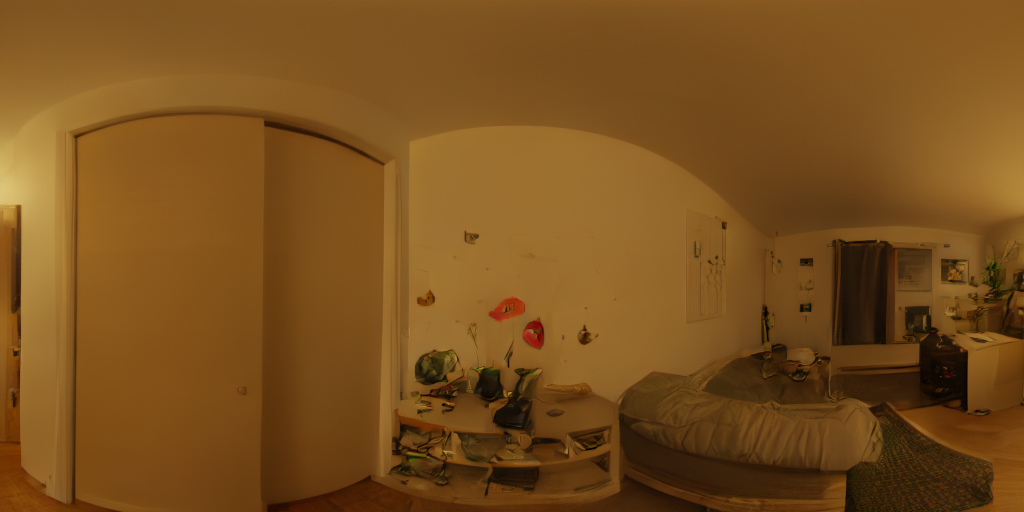} &	
        \includegraphics[width=\customwidth, keepaspectratio]{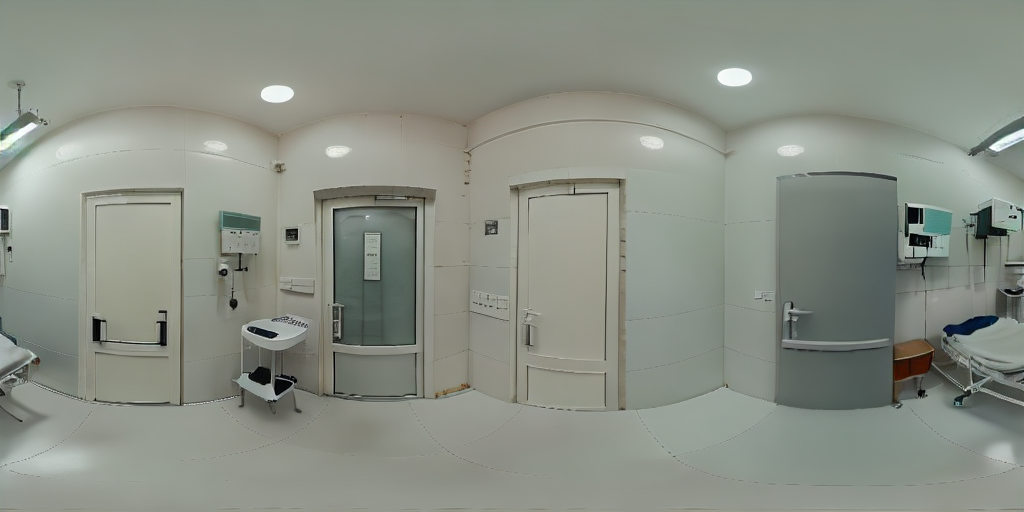} &
    	\includegraphics[width=\customwidth, keepaspectratio]{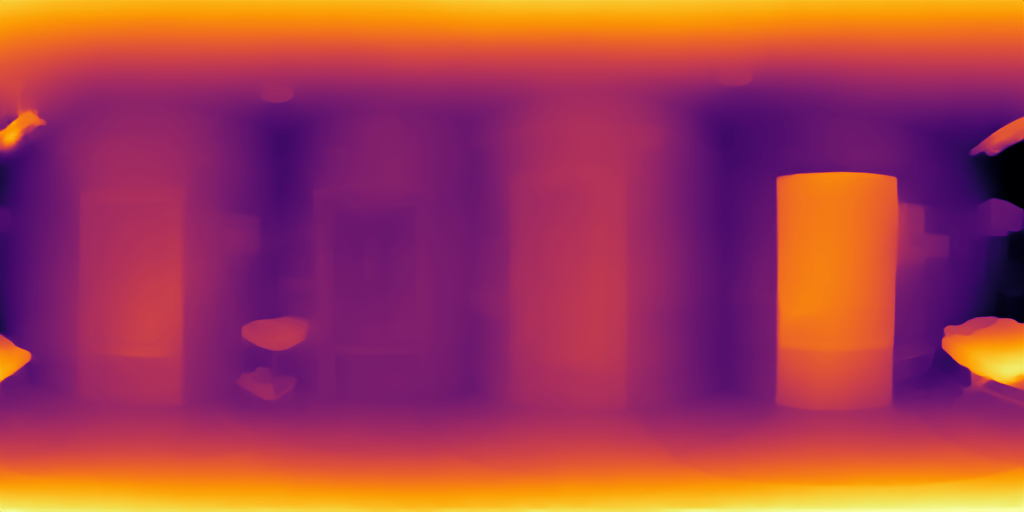} &
    	\includegraphics[width=\customwidth, keepaspectratio]{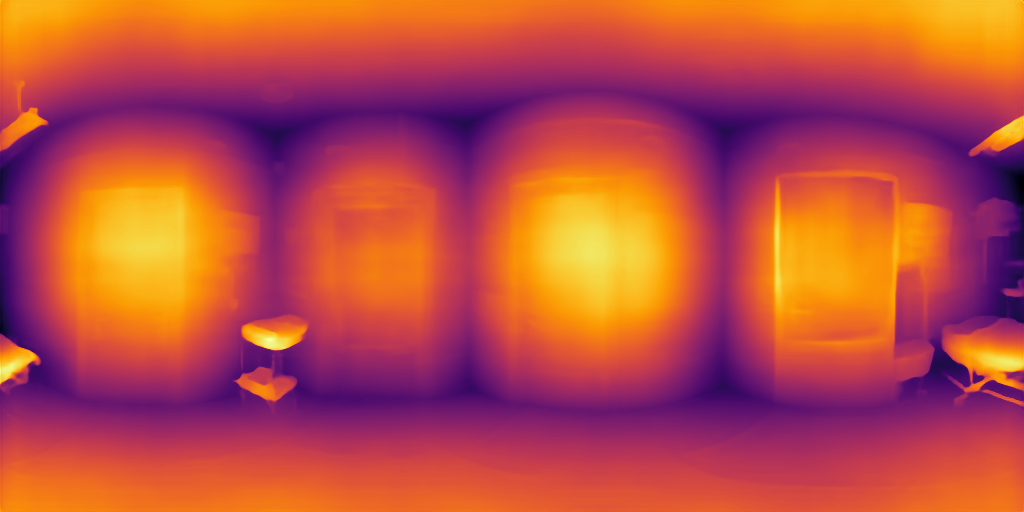}
        \\
      \end{tabular}
      \vspace{-12pt}
      \caption{Qualitative comparison of text-to-panoramic RGBD generation at 512x1024. Images are compared with Text2light~\cite{chen2023text2light}. Depth maps are compared to Joint\_3D60\cite{yun2022improving}. Captions: top--"a 360 view of a field with a few buildings in the distance," middle--"a 360 view of a city street with a bridge," bottom--"a 360 view of a hospital room."}
    \label{fig:pano-rgbd-visual}
  \end{minipage}
  \hspace{.03\linewidth}
  \begin{minipage}{.26\linewidth}
    \centering
    \vspace{6pt}
    \includegraphics[width=\linewidth, keepaspectratio]{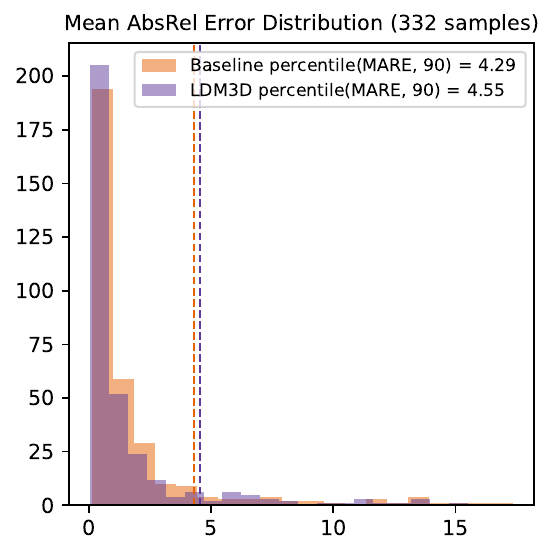}
    \vspace{-18pt}
    \caption{Error distribution across evaluated samples shows a long tail. This motivates removing outliers with error above the 90th percentile.}
    \label{fig:pano-mare-distribution}
  \end{minipage}
\end{figure}

For image quality assessment, we utilize Frechet Inception Distance (FID), Inception Score (IS), and CLIP similarity; these metrics are summarized in Table~\ref{pano-image-metrics}. LDM3D-pano achieves higher FID, and comparable IS and CLIPsim compared to Text2light. LDM3D-pano's higher FID may be due to its deficiency in local awareness and the absence of training in patch-based semantic coherence; focusing on the overall, global context of the given text, potentially at the expense of finer, localized details. Nonetheless, by leveraging extensive text-to-image pretraining from~\cite{rombach2022high, ldm3d}, LDM3D-pano has the capacity to generate a diverse range of images, as is reflected by its marginally higher IS and CLIPsim scores, and in visualized samples in Figure \ref{fig:pano-rgbd-visual}.

\label{pano-depth-evaluation}
\textbf{Depth evaluation.} For panoramic depth evaluation, we compare LDM3D-pano to a baseline monocular panorama depth estimation model: Joint\_3D60\_Fres model~\cite{yun2022improving}. Since diffused RGBD outputs have no ground truth depth available, we use DPT-BEiT-L-512~\cite{birkl2023midas} depth as reference.

The evaluated depth maps and the reference depth are all in disparity space and thus non-metric. We primarily consider the mean absolute relative error (MARE). We fit depth estimates to the reference via least-squares over 500 randomly sampled points. This aims to rescale and reshift the depth estimates to be more closely aligned with the reference. We then compute the MARE and summarize results Table~\ref{pano-depth-metrics}. As explained in Figure~\ref{fig:pano-mare-distribution}, we also report the MARE computed while excluding outlier samples where error exceeds the 90th percentile. In both cases, LDM3D-pano achieves lower MARE than the baseline panoramic depth estimation model.

\begin{table} [h!]
  \begin{minipage}{.525\linewidth}
  \caption{Text-to-pano image metrics at 512x1024, evaluated on 332 samples from our validation set.}
  \vspace{-6pt}
  \label{pano-image-metrics}
  \centering
  \begin{tabular}{@{}*3{l@{\hspace{3mm}}}l@{}}
    \toprule
    Method     & \multicolumn{1}{c}{FID $\downarrow$}    & \multicolumn{1}{c}{IS $\uparrow$}    & \multicolumn{1}{c}{CLIPsim $\uparrow$}    \\
    \midrule
    Text2light\cite{chen2023text2light} & \textbf{108.30}  & 4.646$\pm$0.27  & 27.083$\pm$3.65     \\
    LDM3D-pano     & 118.07       & \textbf{4.687$\pm$0.50}  & \textbf{27.210$\pm$3.24}   \\
    \bottomrule
  \end{tabular}
  \end{minipage}
  \hspace{.02\linewidth}
  \begin{minipage}{.43\linewidth}
  \caption{Pano depth metrics at 512x1024. Reference depth is from DPT-BEiT-L-512.}
  \vspace{-6pt}
  \label{pano-depth-metrics}
  \centering
  \begin{tabular}{@{}*2{l@{\hspace{3mm}}}l@{}}
    \toprule
    Method & \multicolumn{1}{c}{MARE $\downarrow$} & \multicolumn{1}{c}{$\leq$90\%ile} \\
    \midrule
    Joint\_3D60\cite{yun2022improving} & 1.75$\pm$2.87  & 0.92$\pm$0.87 \\
    LDM3D-pano & \textbf{1.54$\pm$2.55}  & \textbf{0.79$\pm$0.77} \\
    \bottomrule
  \end{tabular}
  \end{minipage}
\end{table}

\subsection{High-resolution RGBD generation}
We evaluate HR-RGBD generation using a subset from ImageNet-Val~\cite{ImageNet} composed of 2243 samples at 512x512 resolution. The LR validation set is constructed via bicubic downscaling of HR-RGBD.

\begin{figure}[!htb]
  \begin{minipage}{1.0\linewidth}
      \renewcommand{\customwidth}{0.16\linewidth}
      \setlength{\tabcolsep}{2pt}
      \centering
      \begin{tabular}{c c c c c c}
        \small{LR Input} & 
        \small{Bicubic} & 
        \small{SDx4\cite{rombach2022high}} & 
        \small{LDMx4\cite{LDMx4}} & 
        \small{SD-superres\cite{superres}} & 
        \small{\textbf{LDM3D-SR-B}}
        \\
        \includegraphics[width=0.165\textwidth, keepaspectratio]{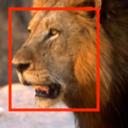} &
        \includegraphics[width=0.143\textwidth, keepaspectratio]{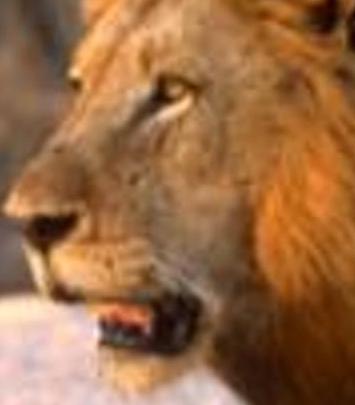} &
        \includegraphics[width=0.143\textwidth, keepaspectratio]{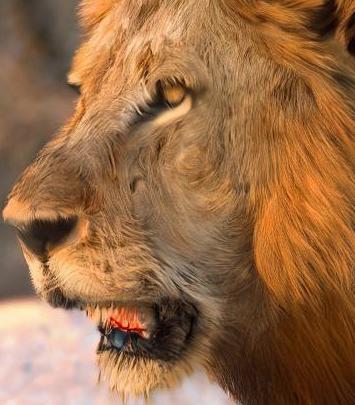} &	       
        \includegraphics[width=0.143\textwidth, keepaspectratio]{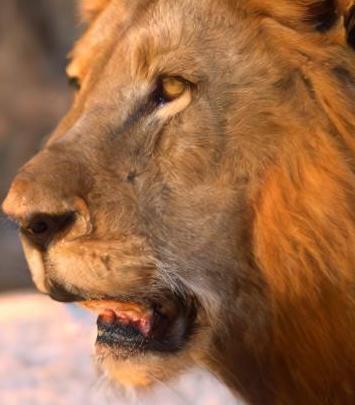} &
    	\includegraphics[width=0.143\textwidth, keepaspectratio]{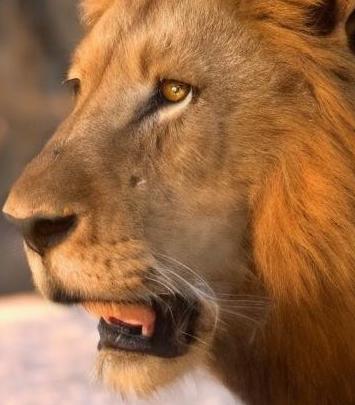} &
    	\includegraphics[width=0.143\textwidth, keepaspectratio]{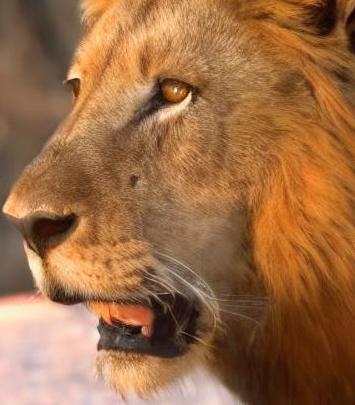}
        \\
      \end{tabular}
      \vspace{-8pt}
      \caption{Qualitative comparison of x4 upscaling. Image sourced from ImageNet-Val.}
      \vspace{4pt}
      \label{fig:hr-upscaling}
  \end{minipage}
  \begin{minipage}{.7\linewidth}
      \renewcommand{\customwidth}{0.2\linewidth}
      \setlength{\tabcolsep}{1pt}
      \centering
      \begin{tabular}{c c c c c}
        \scriptsize{LR Image} &
        \multicolumn{2}{c}{\scriptsize{\textbf{RGBD from LDM3D-SR-B}}} &
        \scriptsize{Bicubic Baseline} &
        \scriptsize{Reference Depth} 
        \\
        \vspace{-1pt}
        \includegraphics[width=\customwidth, keepaspectratio]{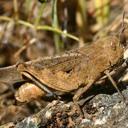} &
        \includegraphics[width=\customwidth, keepaspectratio]{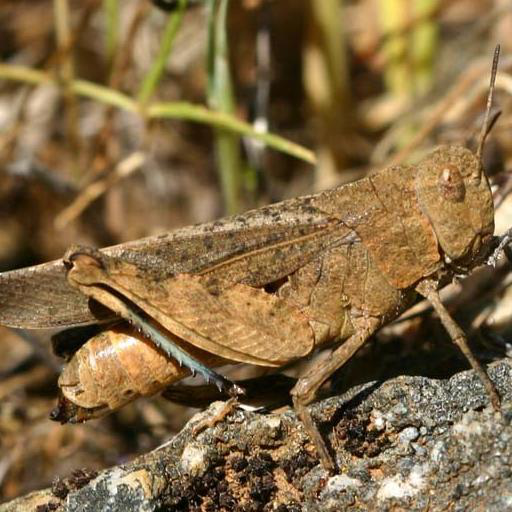} &	    
        \includegraphics[width=\customwidth, keepaspectratio]{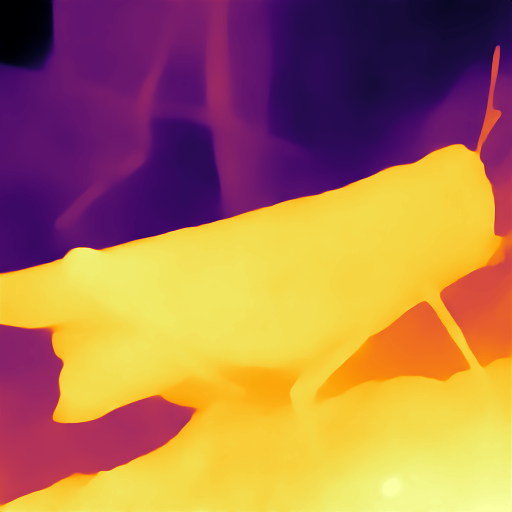} &
        \includegraphics[width=\customwidth, keepaspectratio]{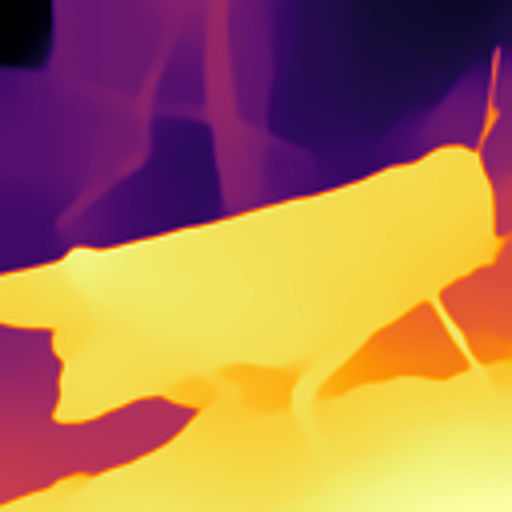} &
    	\includegraphics[width=\customwidth, keepaspectratio]{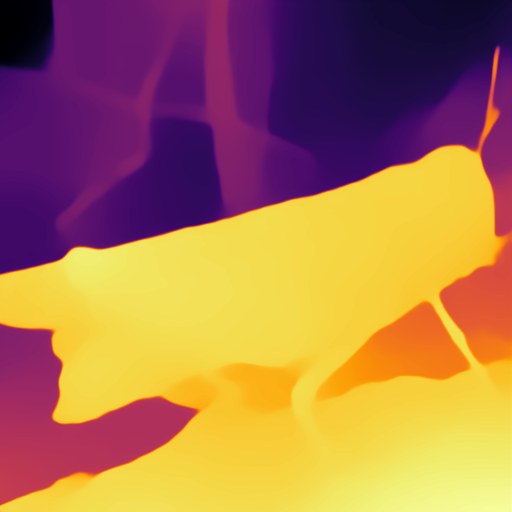}
        \\
        \includegraphics[width=\customwidth, keepaspectratio]{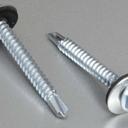} &
        \includegraphics[width=\customwidth, keepaspectratio]{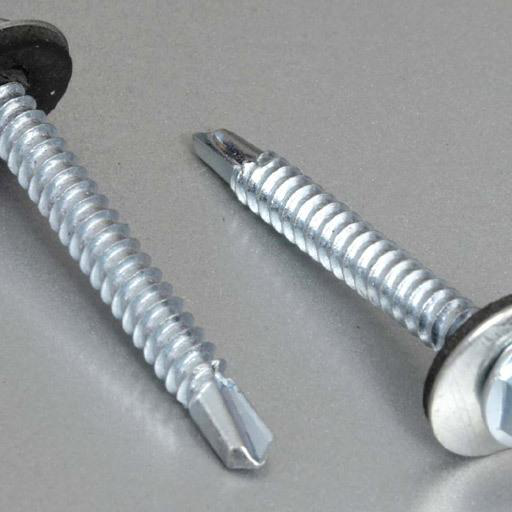} &	     
    	\includegraphics[width=\customwidth, keepaspectratio]{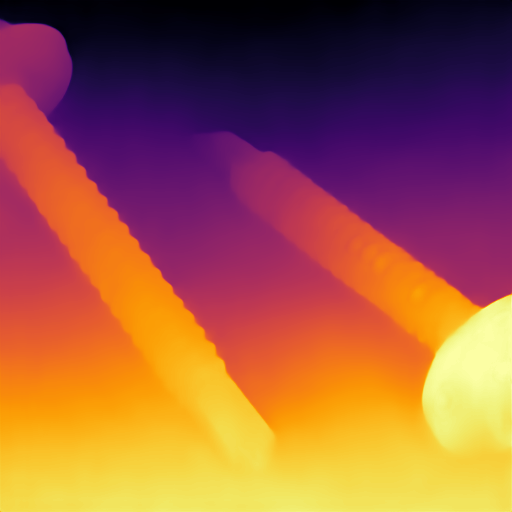} &
    	\includegraphics[width=\customwidth, keepaspectratio]{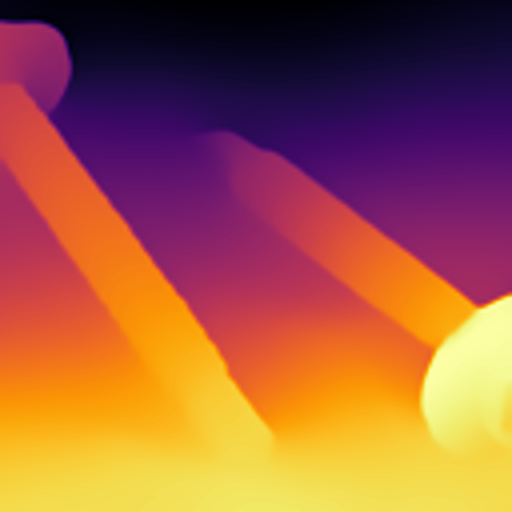} &
    	\includegraphics[width=\customwidth, keepaspectratio]{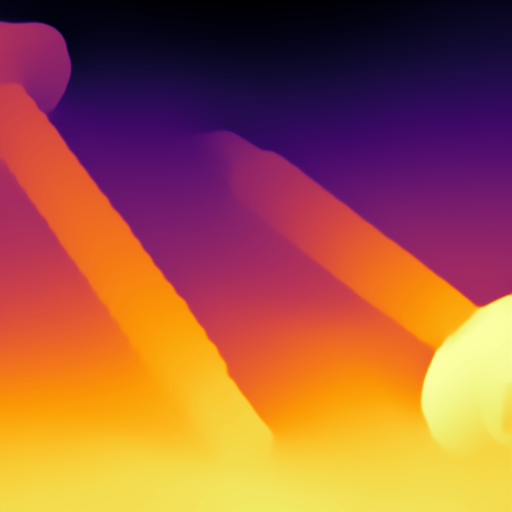}
        \\
      \end{tabular}
      \vspace{-12pt}
      \caption{Additional visualization of x4 RGBD upscaling with LDM3D-SR-B. Depth maps are compared to a bicubic regression baseline against reference depth obtained using DPT-BEiT-L-512.}
      \label{fig:hr-depth-visualization}
  \end{minipage}
  \hspace{.03\linewidth}
  \begin{minipage}{.26\linewidth}
    \centering
    \vspace{3pt}
    \includegraphics[width=\linewidth, keepaspectratio]{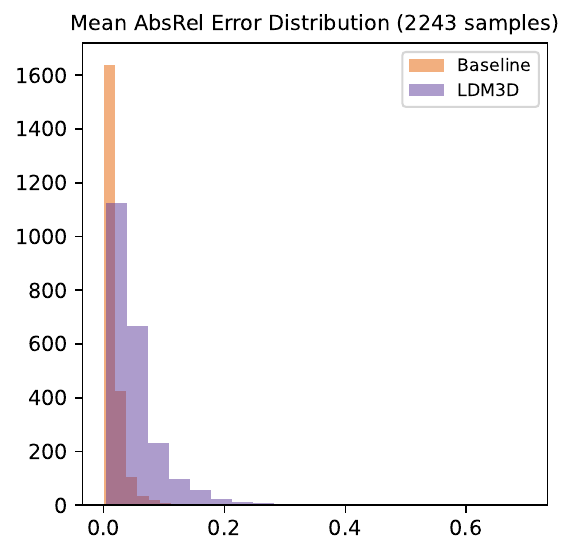}
    \vspace{-18pt}
    \caption{Since there is no long tail in the error distribution across evaluated ImageNet samples, no outlier removal is performed.}
    \label{fig:hr-mare-distribution}
  \end{minipage}
\end{figure}

\textbf{Image evaluation.} In line with previous studies, we use reconstruction FID, IS, Peak Signal-to-Noise Ratio (PSNR), and Structural Similarity Index Measure (SSIM) to evaluate the quality of HR images; these metrics are summarized in Table \ref{hr-eval-metrics}. We compare LDM3D-SR against bicubic regression, SDx4 super-resolution based on SDv2 \cite{rombach2022high}, LDMx4 \cite{LDMx4} and SD-superres based on SDv1 \cite{superres}. The latter is the model from which we fine-tune LDM3D-SR. Our findings reveal that LDM3D-SR achieves the best FID and the second-best IS after SDx4. Conversely, bicubic regression attains the highest scores on PSNR and SSIM. However, these two metrics tend to prefer blurriness over misaligned high-frequency details \cite{saharia2021image} and often contradict human perception \cite{Perceptualmetric}; this is supported by the significantly lower FID score of bicubic regression as well as by the visualization in Figure \ref{fig:hr-upscaling}. 

In comparing LDM3D-SR-D, LDM3D-SR-O, and LDM3D-SR-B, Table \ref{hr-eval-metrics} also presents an ablation study on the optimal depth preprocessing method outlined in \ref{ldm3d-hr-finetuned}. Results reveal that employing bicubic degradation of the initial depth map is closely aligned with utilizing the original HR depth map as conditioning. Conversely, utilizing a depth map calculated from the degraded image yields inferior results, likely due to low quality depth maps.

\textbf{Depth evaluation.} Our depth evaluation protocol for LDM3D-SR closely follows that used for LDM3D-pano, as described in~\ref{pano-depth-evaluation}. Our baseline here is bicubic regression on depth. Table~\ref{hr-eval-metrics} reports the MARE for bicubic regression and our LDM3D-SR methods. The error in bicubically-regressed depth maps is predominantly along object boundaries that become blurred upon bicubic interpolation; since edges account for a small fraction of scene content, the MARE for bicubic regression is particularly low. Amongst the LDM3D-SR methods, -D exhibits the highest MARE while -O and -B both exhibit a lower MARE. We show additional visualizations of RGBD upscaling in Figure~\ref{fig:hr-depth-visualization}, where we observe high-resolution features in both the images and depth maps (the wings and antennae of the grasshopper, the threads of the screw). Lastly, Figure~\ref{fig:hr-mare-distribution} does not indicate the presence of a tail in the error distribution, so we do not perform any outlier removal in this evaluation.

\begin{table} [h!]
  \caption{x4 upscaling from 128x128 to 512x512, evaluated on 2243 samples from ImageNet-Val}
  \vspace{-6pt}
  \label{hr-eval-metrics}
  \centering
  \begin{tabular}{@{}l*5{@{\hspace{4mm}}c}@{}}
    \toprule
    Method     & \multicolumn{1}{c}{FID $\downarrow$}    & \multicolumn{1}{c}{IS $\uparrow$}    & \multicolumn{1}{c}{PSNR $\uparrow$}    & \multicolumn{1}{c}{SSIM $\uparrow$} &  \multicolumn{1}{c}{Depth MARE $\downarrow$} \\
    \midrule
    Regression, bicubic & 24.686  & 60.135$\pm$4.16  & \textbf{26.424$\pm$3.98}  & \textbf{0.716$\pm$0.13} & \textbf{0.0153$\pm$0.0189}\\
    SDx4\cite{rombach2022high}     & 15.865 &  \textbf{61.103$\pm$3.48}  & 24.528$\pm$3.63  & 0.631$\pm$0.15 & N/A \\
    LDMx4\cite{LDMx4}     & 15.245 &60.060$\pm$3.88  &  \underline{25.511$\pm$3.94}  & \underline{0.686$\pm$0.16} & N/A\\
    SD-superres\cite{superres}     & 15.254      & 59.789$\pm$3.53  & 23.878$\pm$3.28  & 0.642$\pm$0.15  & N/A \\
    LDM3D-SR-D     & 15.522       &  59.736$\pm$3.37  & 24.113$\pm$3.54  & 0.659$\pm$0.16  & 0.0753$\pm$0.0734 \\
    LDM3D-SR-O     & \underline{14.793}       & 60.260$\pm$3.53  & 24.498$\pm$3.59  & 0.665$\pm$0.16  & \underline{0.0530$\pm$0.0496}\\
    LDM3D-SR-B     & \textbf{14.705}       &  \underline{60.371$\pm$3.56}  & 24.479$\pm$3.58  & 0.665$\pm$0.48 & 0.0537$\pm$0.0506\\
    \bottomrule
  \end{tabular}
\end{table}

\vspace{-6pt}

\section{Conclusion}
We introduce LDM3D-pano and LDM3D-SR for 3D VR applications. LDM3D-pano competes with panorama-specialized models by generating diverse high-quality panoramic images jointly with panoramic depth. LDM3D-SR focuses on RGBD upscaling, outperforming related image upscaling methods while also generating high-resolution depth maps. Future work could combine these domains to generate high-resolution panorama RGBD to further enhance immersive VR experiences.

{
\small
\bibliographystyle{plain}
\bibliography{refs}
}


\end{document}